\documentclass{article}
\usepackage{graphicx} 
\usepackage{amsmath,amssymb,amsthm}
\usepackage[colorlinks=true, allcolors=blue]{hyperref}
\usepackage{algorithm}
\usepackage{algpseudocode}
\usepackage{bbm}
\usepackage[authoryear,round]{natbib}
\usepackage{xcolor}
\usepackage[affil-it]{authblk}
\usepackage{tikz}
\usetikzlibrary{positioning}
\usepackage{subcaption}

\newtheorem{theorem}{Theorem}

\newtheorem{lemma}[theorem]{Lemma}
\newtheorem{proposition}{Proposition}
\theoremstyle{definition}
\newtheorem{definition}{Definition}
\newtheorem{assumption}{Assumption}
\theoremstyle{remark}
\newtheorem{remark}{Remark}

\newcommand{\R}{\mathbb{R}}
\newcommand{\E}{\mathbb{E}}

\newcommand{\var}{\text{Var}}
\newcommand{\ch}{\operatorname{Ch}}

\newcommand{\gap}{\operatorname{Gap}}

\title{Spectral gap of Metropolis-within-Gibbs under log-concavity}
\author{Cecilia Secchi\thanks{Bocconi University, Department of Decision Sciences, Milan, Italy.}\;  and Giacomo Zanella\thanks{Bocconi University, Department of Decision Sciences and BIDSA, Milan, Italy.\\ GZ acknowledges support from the European Research Council (ERC), through StG ``PrSc-HDBayLe''
grant ID 101076564.}}
\date{}

\begin{document}

\maketitle

\begin{abstract}

The Metropolis-within-Gibbs (MwG) algorithm is a widely used Markov chain Monte Carlo method for sampling from high-dimensional distributions when exact conditional sampling is intractable. We study MwG with Random Walk Metropolis (RWM) updates, whose proposal variances are uniformly comparable to the corresponding conditional variances.
Assuming the target $\pi$ is a $d$-dimensional log-concave distribution with condition number $\kappa$, we establish a spectral gap lower bound of order $\Omega((\kappa d)^{-1})$ for the random-scan version of MwG, improving on the previously available $\Omega((\kappa^2 d)^{-1})$ bound. 
This is obtained by developing sharp estimates of the conductance of one-dimensional RWM kernels, which may be of independent interest. 
The result shows that MwG can mix substantially faster under the stated uniform tuning condition and that its mixing performance is just a constant factor worse than that of the exact Gibbs sampler, thus providing theoretical support to previously observed empirical behavior.
\end{abstract}

\section{Introduction}

The random-scan Gibbs sampler (GS) \citep{casella1992explaining} is a classical and popular coordinate-wise Markov chain Monte Carlo (MCMC) algorithm \citep{brooks2011handbook}, which can be used to sample from a multivariate probability distribution $\pi$ over $\mathbb{R}^d$. At each iteration, it randomly selects a coordinate $m$ and updates it by sampling from the corresponding conditional distribution of $\pi$, while keeping all other coordinates fixed. 
Thus, to be implementable, the GS requires sampling from one-dimensional conditional distributions, which is a much easier task than sampling directly from the full target $\pi$; however, in many applications such as non-conjugate Bayesian models, one-dimensional conditional distributions may themselves be intractable.
The Metropolis-within-Gibbs (MwG) algorithm overcomes this limitation by replacing exact updates with Metropolis-Hastings (MH) steps \citep{chib1995understanding}, or more generally, Markov updates that have the desired conditionals as invariant distributions. 
A popular choice is the Random Walk Metropolis (RWM) update, in which the selected coordinate is perturbed with zero-mean Gaussian noise and the new state is accepted or rejected according to the MH rule.\\

We study the convergence speed of the MwG algorithm with RWM updates, under the assumption that the target distribution $\pi$ is log-concave with condition number $\kappa$. Existing convergence bounds \citep{qin2025spectral,ascolani2024entropy} suggest that MwG may be slower than the exact GS by a factor $\kappa$ due to the penalty introduced by the Metropolis–Hastings accept-reject step. 
In this work, we show that the extra dependence on $\kappa$ can be eliminated: 
if the proposal variances are optimally tuned, meaning that they are state-dependent and uniformly comparable to the corresponding conditional variances, MwG is slower than GS by at most a constant factor, independent of both $\pi$ and the dimension $d$. Consequently, both algorithms admit a mixing time upper bound of order $O(\kappa d \log(1/\varepsilon))$.
These results are consistent with empirical evidence showing limited (sometimes negligible) slowdown in mixing when moving from GS to a properly tuned MwG, see, e.g., the numerical simulations in \citet{ascolani2026scalability}, \citet{luu2024gibbs} and in Section \ref{sec:application} below, and 
provide theoretical support for MwG as a feasible and efficient alternative to GS for non-conjugate models. 
\\

In some models, evaluating a one-dimensional conditional distribution is $d$ times computationally cheaper than evaluating $\pi$ or its gradient.
Under the uniform tuning condition described above, our spectral-gap bound implies that MwG, like GS, requires a total computational cost equivalent to $\kappa$ full target evaluations to obtain an approximate sample, independently of the dimension $d$.
This compares favorably with classical gradient-based methods, the cost of which grows polynomially with $d$.
For more discussion and explicit theoretical comparisons, see \citet[Section~4.2]{ascolani2024entropy}.

Additionally, the result implies that, in the context of MwG with one-dimensional conditionals, uniformly well-tuned RWM updates attain the optimal worst-case spectral-gap dependence on $\kappa$ and $d$; see Section \ref{section3} for more details. Consequently, more sophisticated reversible conditional updating schemes, such as MALA-within-Gibbs or HMC-within-Gibbs \citep{tong2020mala}, may improve multiplicative constants and practical performance, but are not expected to improve the worst-case dependence on $d$ and $\kappa^*$ among MwG algorithms.\\

We analyze the convergence rate of MwG through its spectral gap, leveraging the following decomposition inequalities that combine results from \cite{qin2025spectral} and \cite{ascolani2024entropy}:
\begin{align}
    \gap(P^\text{MwG})&\geq \left[\inf_{z}\gap(P^z)\right]\cdot\gap(P^\text{GS})\label{spectralgapdec}\\
    &\geq \left[\inf_{z}\gap(P^z)\right]\cdot\frac{1}{\kappa d},
\end{align}
where $P^{\text{MwG}}$ and $P^{\text{GS}}$ are the Markov transition kernels of MwG and GS respectively, and $\{P^z\}$ denotes the family of one-dimensional conditional kernels defining MwG (see Section \ref{section3} for a formal definition).

Exploiting classical isoperimetric inequalities for log-concave distributions, the analysis of the spectral gaps of these one-dimensional kernels reduces the problem to finding a lower bound on the RWM acceptance rate (Theorem \ref{result1}). 
To achieve this, we demonstrate that tuning the proposal variance to match the order of the corresponding conditional variance of $\pi$ ensures that the acceptance probability remains uniformly bounded below by a positive constant (Proposition \ref{boundacceptance}), thus obtaining our desired result.\\

The convergence properties of MwG and its variants, also referred to as hybrid GS, have received considerable attention over the past decades. 
Early works established sufficient conditions for the geometric ergodicity of MwG, including \cite{roberts1998two}, \cite{qin2022convergence} and \cite{fort2003geometric}.
More recently, further 
progress was made through a more quantitative spectral analysis: 
specifically, \citet{qin2025spectral} derived the spectral gap comparison bound in \eqref{spectralgapdec},
 \citet{ascolani2026scalability} developed analogous s-conductance bounds, and 
\cite{qin2025spectral2} generalized and unified the theoretical framework under which the spectral gap decomposition \eqref{spectralgapdec} holds.

Related work on coordinate-wise Langevin Monte Carlo methods for sampling from log-concave distributions includes that of \cite{tong2020mala}, \cite{ding2021randomU} and \cite{ding2021random}.\\

We will work on 
$(\mathbb{R}^d, \mathcal{B}(\mathbb{R}^d))$, where $\mathcal{B}(\mathbb{R}^d)$ denotes the Borel $\sigma$-algebra.
We denote by $\mathcal{P}(\R^d)$ the space of probability measures on $(\R^d,\mathcal{B}(\R^d))$ absolutely continuous with respect to Lebesgue measure. 

\section{Conductance lower bound for RWM}

In this section, we establish a lower bound on the conductance of the Markov transition kernel associated with the RWM algorithm (Theorem \ref{result1}). 
Crucially, when $d=1$ our lower bound is uniformly greater than a positive constant regardless of the value of $\kappa$.

Our approach follows a classical framework for lower bounding the conductance of reversible Markov chains. It combines a one-step overlap condition on the transition kernel with an isoperimetric inequality for the target distribution. A pivotal step in the analysis is to bound the acceptance rate, which we prove to remain bounded away from zero for all one-dimensional log-concave target distributions when the proposal variance is of the same order as the target variance.

\subsection{Main results}
We begin by introducing the necessary notation and definitions required to state the main results.
Let $\pi\in\mathcal{P}(\R^d)$ be the target distribution from which we aim to sample. A Markov chain is specified by its transition kernel $P:\R^d\times\mathcal{B}(\R^d)\rightarrow[0,1]$. 
 $P$ is said to be $\pi$-\textit{reversible} if 
 \[\pi(dx)P(x,dy)=\pi(dy)P(y,dx),\]
 where the above is an equality of measures defined on $\R^d\times \R^d$ endowed with its Borel $\sigma$-algebra. 
The kernel $P$ is $\pi$-\textit{invariant} if, for any
  $B\in\mathcal{B}(\R^d)$, 
 \[
 \int_{\R^d}P(x,B)\pi(dx)=\pi(B).
 \] 
 Note that $\pi$-reversibility implies $\pi$-invariance, but the converse is generally not true.

The Metropolis-Hastings (MH) kernel is a $\pi$-reversible kernel with structure 
\[
P(x,dy):=\alpha(x,y)Q(x,dy)+r(x)\delta_x(dy),
\]
where $Q:\R^d\times\mathcal{B}(\R^d)\rightarrow[0,1]$ is the proposal kernel and $\alpha(x,y)$ is the acceptance probability for the proposal $y$ given the current point $x$
defined as 
\[
\alpha(x,y):=\min\left\{1,\frac{\pi(dy)Q(y,dx)}{\pi(dx)Q(x,dy)}\right\}.
\]
The average acceptance rate at $x$ is $\alpha(x):=\int \alpha(x,y) Q(x,dy)$ and its complement is $r(x):=1-\alpha(x)$. 
We denote the infimum of the acceptance rate as 
\[
\alpha_0:=\inf_{x\in\R^d} \alpha(x).
\]
For the RWM kernel, the proposal is symmetric, with distribution of the form $Q(x,\cdot)\sim\mathcal{N}(x,\sigma^2I_d)$, for some $\sigma^2>0$. In this case, the acceptance probability simplifies to
\[
\alpha(x,y)=\min\left\{1,\frac{\pi(y)}{\pi(x)}\right\},
\]
where $\pi(\cdot)$ denotes the density of the target distribution with respect to the Lebesgue measure. 
We summarize the considered framework in the following assumption.
\begin{assumption}\label{assA}
$P$ is a RWM kernel with $Q(x,\cdot)\sim\mathcal{N}(x,\sigma^2I_d)$ and $\pi\in\mathcal{P}(\R^d)$ has density $\pi(x) \propto \exp(-U(x))$, where $U: \mathbb{R}^d \to \mathbb{R}$ is a continuously differentiable strictly convex potential. 
\end{assumption}

\begin{proposition}\label{prop:rwm-positive}
Under Assumption \ref{assA}, the RWM kernel $P$ is
$\pi$-reversible and positive on $L^2(\pi)$, that is 
$\langle f,Pf\rangle_\pi\geq 0$ for every $f\in L^2(\pi)$,
where for $f,g\in L^2(\pi)$, $\langle f,g\rangle_\pi
:=
\int f(x)g(x)\,\pi(dx)$.
\end{proposition}
\begin{proof}
Reversibility follows from the Metropolis-Hastings construction.
Positivity follows from
\citet[Lemma~3.1 and Remark~3.2]{baxendale2005renewal}, as the Gaussian proposal kernel $Q(x,\cdot)=\mathcal{N}(x,\sigma^2I_d)$
is the two-step kernel associated with the symmetric Gaussian kernel
$ R(x,\cdot)= \mathcal{N}\left(x,\sigma^2/2 \,I_d\right)$.
\end{proof}

Our analysis relies on the notion of conductance, defined as follows.
\begin{definition}
    The \textit{conductance}  of a $\pi$-invariant Markov kernel $P$ is 
\[
\Phi(P)=\inf_{\substack{A\in\mathcal{B}(\R^d), \\ \pi(A)\in(0,1)}}\frac{\int_A P(u,A^c)\pi(du)}{\min\{\pi(A),\pi(A^c)\}}.
\]
\end{definition}
Conductance quantifies the probability that a one-step transition of the Markov chain exits a given set $A$, with $\pi(A)\le1/2,$ when the chain is initialized according to the normalized conditional distribution $\pi(\cdot\mid A)=\pi(\cdot\,\cap A)/\pi(A)$. High conductance indicates the absence of bottlenecks and guarantees rapid mixing, as we will see below.

We are now ready to state the main result of this section.

\begin{theorem}\label{result1} Under Assumption \ref{assA} we have:
\begin{enumerate}
    \item[(a)] 
given $K=6\sqrt{3}$ and $X\sim\pi$,
\begin{equation*}\label{lowerboundphu}
\Phi(P)\geq \min\left\{\frac{\alpha_0}{8},\frac{\alpha_0^2\,\sigma}{16\,K\sqrt{\E_\pi[\|X-\E_\pi [X]\|^2]}}\right\}, 
\end{equation*}   
    \item[(b)] if in addition $d=1$ and $\sigma^2=c\,\var(\pi)$ for some $c>0$, then
\[
\Phi(P)\geq \min\left\{\frac{b(c)}{8},\frac{b(c)^2\sqrt{c}}{16\,K}\right\}=:k(c),
\]
where $b(c)=\phi_c(3/2)\frac{e^{-2.6}}{2}$ and $ \phi_c$ is the density of $\mathcal{N}(0,c) $.
\end{enumerate}
\end{theorem}

Notably, restricting to $d=1$ enables us to work under minimal assumptions on the potential $U$. Specifically, we do not require $U$ to be strongly convex or smooth, conditions which are typically needed in higher-dimensional analyses. Consequently, when applied to strongly log-concave and smooth targets, the bounds in Theorem \ref{result1}(b) remain independent of the condition number $\kappa$.

Having a bound on the conductance also allows us to control the spectral gap of $P$, defined as follows.
\begin{definition}
Let $P$ be a positive, $\pi$-reversible Markov kernel. Its \emph{spectral gap} is 
\[
\gap(P)=\inf_{\substack{f\in L^2_0(\pi)\\ f\neq 0}}
\frac{\langle f,(I-P)f\rangle_\pi}{\|f\|_\pi^{2}}\in[0,1],
\qquad
L^2_0(\pi)=\Bigl\{f\in L^2(\pi):\textstyle\int f\,d\pi=0\Bigr\}.
\]
where $Pf(x)=\int f(y) P(x,dy)$ and $\|f\|_\pi^2=\langle f,f\rangle_\pi$.
\end{definition}
    
Under positivity, the spectral gap is related to the operator norm of $P$ through the identity
 \[
 \gap(P)=1-\|P\|_\pi,\]
 where
$\|P\|_\pi:=\sup\left\{ \|Pf\|_\pi/\|f\|_\pi:\, f\in L^2(\pi),\,\E_\pi(f)=0\right\}$ \citep[§14, Corollary 5.1]{helmberg2008introduction}. Hence $\gap(P)$ is directly connected to the convergence rate of the Markov kernel, as it quantifies how fast $P^tf$ converges to $0$ for $f\in L_0^2(\pi)$. Specifically, for any $f\in L^2_0(\pi)$, it holds
\begin{equation}\label{conv}
    \|P f\|_\pi\leq (1-\gap(P))\|f\|_\pi.
\end{equation}
The conductance and spectral gap are connected via Cheeger's inequality \citep[Lemma~5]{andrieu2024explicit}, \citep[Theorem~3.5]{lawler1988bounds}:
\begin{equation}\label{cheeger}
    \frac{1}{2}\Phi(P)^2\leq\gap(P)\leq 2\Phi(P).
\end{equation}
Combining this with Theorem \ref{result1}(b), we obtain 
\begin{equation}\label{lowerboundspectral}
    \gap(P)\geq \frac{k(c)^2}{2}.
\end{equation}

\subsection{Proof of Theorem \ref{result1}(a)}
The proof of Theorem \ref{result1}(a) follows from a general lower bound of the conductance under isoperimetric and overlap conditions. We begin by introducing the relevant definitions.

For measurable sets $A, B \in \mathcal{B}(\mathbb{R}^d)$, we define the set distance as $d(A,B):=\inf\{\|x-y\|\,:\,x\in A,\, y\in B\}$.

\begin{definition}\label{cheeger1}
For $A\in\mathcal{B}(\R^d)$, the $r$-neighborhood $A^r$ is defined as $A^r:=\{x\in \R^d\;:\;d(x,A)<r\}$. The \textit{boundary measure} of $A$ with respect to a probability measure $\pi\in\mathcal{P}(\R^d)$ is defined as 
\[
\pi^+(A):=\liminf_{r\rightarrow 0^+}\frac{\pi(A^r)-\pi(A)}{r}.
\]
We say that $\pi$ satisfies a \textit{Cheeger isoperimetric inequality} with constant $\ch(\pi)>0$ if for all $A\in\mathcal{B}( \R^d)$
\[
\pi^+(A)\geq \frac{1}{\ch(\pi)}\min\{\pi(A),\pi(A^c)\}.
\]
The smallest such constant $\ch(\pi)$ is called the \textit{Cheeger constant} of $\pi$.
\end{definition}
This constant captures the presence of bottlenecks in the $\pi$-weighted space $\R^d$, as it provides a lower bound on the surface area-to-volume ratio $\pi^+(A)/\pi(A)$ over all possible two-partitions of the space. 

Recalling that the total variation distance between two probability measures $\mu,\nu\in\mathcal{P}(\R^d)$ is defined as $d_{TV}(\mu,\nu)=\sup_{A\in\mathcal{B}(\R^d)}|\mu(A)-\nu(A)|$, we can state the following well-known lemma, a proof of which can be found, for example, in \citet[Lemma 5.4]{lee2024fast}.
\begin{lemma} \label{2cond}
    Let $P$ be a Markov transition kernel on $\R^d$ with stationary distribution $\pi$. Suppose:
    \begin{enumerate}
        \item (Isoperimetry) $\pi$ satisfies a Cheeger isoperimetric inequality with constant $\ch(\pi)>0$;
        \item (One-step overlap) there exists $\delta>0$ such that for any points $x,y\in\R^d$ with $\|x-y\|\leq \delta$, it holds that $d_{TV}(P(x,\cdot),P(y,\cdot))\leq 1-h$ for some $0<h<1$.
    \end{enumerate}
    Then the conductance of $P$ satisfies
    \[\Phi(P)\geq \min\left\{\frac{h}{4},\frac{\delta\,h}{8\ch(\pi)}\right\}.\]
\end{lemma}
While conductance is a global property that depends on all subsets of the state space, this lemma provides a more tractable criterion in terms of an intrinsic property of the target measure together with a local property of the transition kernel.

We now verify that the RWM kernel under Assumption \ref{assA} satisfies the conditions of Lemma \ref{2cond}, combining previous results from the literature.
\begin{lemma}\label{isoperimetric-log-concave} \citep[Theorem 1.2]{bobkov1999isoperimetric}
    Let $\pi\in\mathcal{P}(\R^d)$ be log-concave. Then
\begin{equation}
    \frac{1}{\ch(\pi)}\geq\frac{1}{K\sqrt{\E[\|X-\E [X]\|^2]}},
\end{equation}
where $K=6\sqrt{3}$ and $X\sim\pi$. 
\end{lemma} 
\begin{lemma}\label{overlap} \citep[Lemma 38]{andrieu2024explicit}
    The RWM kernel with proposal variance $\sigma^2$ satisfies the one-step overlap condition with $\delta=\alpha_0\,\sigma$ and $h=\frac{1}{2}\,\alpha_0$. 
\end{lemma}
\begin{proof}[Proof of Theorem \ref{result1}(a)]
The desired result is obtained by combining Lemmas \ref{isoperimetric-log-concave} and \ref{overlap} in Lemma \ref{2cond}.
\end{proof}

\subsection{Proof of Theorem \ref{result1}(b)}
The proof of Theorem \ref{result1}(b) relies on establishing a lower bound on the acceptance rate $\alpha_0$ for the RWM algorithm in the one-dimensional case.

\begin{proposition}\label{boundacceptance}
Let Assumption \ref{assA} hold with $d=1$ and $\sigma^2=c\,\var(\pi)$ for $c>0$. 
Then 
\[
\alpha(x) = \E \left[ \min\left\{1,\frac{\pi(x+Z)}{\pi(x)} \right\} \right] \geq b(c),
\]  
where $ Z \sim \mathcal{N}(0,\sigma^2)$ and $b(c)=\phi_c(3/2)\frac{e^{-2.6}}{2}$, with $ \phi_c$ denoting the density of $\mathcal{N}(0,c) $.
\end{proposition}

\begin{proof}
We first reduce to the case $\var(\pi)=1$.
Let $\gamma^2=\var(\pi)$ and define the rescaled density $\bar\pi(u)=\gamma \,\pi(\gamma \,u) $. If $X\sim\pi$, then $X/\gamma\sim\bar\pi$ and $\var(\bar\pi)=1$.
Moreover, since $\sigma^2=c\gamma^2$, if $Z\sim \mathcal{N}(0,\sigma^2)$, then $\bar Z:=Z/\gamma\sim\mathcal{N}(0,c)$, consequently,
\begin{equation*}
\alpha(x)
=
\E\left[
\min\left\{1,\frac{\pi(x+Z)}{\pi(x)}\right\}
\right] 
=
\E\left[
\min\left\{1,
\frac{\bar{\pi}(x/\gamma+\bar Z)}{\bar{\pi}(x/\gamma)}
\right\}
\right].
\end{equation*}
Thus, it is sufficient to prove the result when $\var(\pi)=1$ and
$\sigma^2=c$.

We may also assume that $U$ attains its minimum at $0$ and that $ U(0) = 0 $. 
Indeed, if $x^*$ is a minimizer of $U$, define $\widetilde U(u)=U(u+x^*)-U(x^*)$ and let $\widetilde\pi$ be the corresponding translated density. Then, writing $\alpha_\nu(x)$ for the acceptance probability associated with a target density $\nu$,
\[
\alpha_\pi(x)=\alpha_{\tilde\pi}(x-x^*).
\]
As $U(0)=U'(0)=0$, by Lemma \ref{boundU1}, $\min\{U(-1),U(1)\}\leq 2.6$.
Without loss of generality, we assume that
\begin{equation}\label{boundu1appl}
U(1)\leq 2.6.
\end{equation}
Indeed, if this inequality does not hold, thanks to the Gaussian proposal symmetry, the same argument applies after reflecting about the origin all the intervals considered in the three cases. 

We now proceed to bound the acceptance probability $\alpha(x)$ by distinguishing three cases based on the value of $x$. 
Define the left endpoint of the sublevel set at height $U(1/2)$ by $U^\leftarrow:=\inf\{t\in\R_{\le 0}:U(t)\leq U(1/2)\}$ and set $y:=\max\{U^\leftarrow,-1/2\}\le 0$.

\noindent\textit{Case 1: $x\in[y,1/2]$.} 
For $w\in[1/2,1]$, convexity and the fact that $0$ is a minimizer imply $U(w)\ge U(1/2)\ge U(x)$, hence $\pi(w)\le\pi(x)$ and therefore
\begin{align*}
    \alpha(x)
    &\ge 
    \int_{-x+1/2}^{-x+1}\min\left\{1,\frac{\pi(x+z)}{\pi(x)}\right\}\phi_c(z)\,dz \\
    &=
    \int_{1/2}^{1} e^{U(x)-U(w)}\phi_c(w-x)\,dw.
\end{align*}
Since $U(x)\ge U(0)$, we have $e^{U(x)-U(w)}\geq e^{-U(w)}$. Moreover, for $w\in[1/2,1]$, convexity gives $U(w)\le U(1)$, while $x\in[y,1/2]\subseteq [-1/2,1/2]$ implies $0\le w-x\le \frac32$.
As $\phi_c$ is nonincreasing on $[0,\infty)$, we obtain
\begin{align*}
    \alpha(x)
    &\ge
    \int_{1/2}^1 e^{-U(w)}\phi_c(w-x)\,dw\\
    &\ge 
    \frac12 e^{-U(1)}\phi_c(3/2)
    \ge 
    \frac 12 e^{-2.6}\phi_c(3/2),
\end{align*}
where the last inequality follows from \eqref{boundu1appl}.
    
\noindent\textit{Case 2: $x\le-1/2$ or $x>1/2$.}
If $x\le-1/2$, then for every $z\in[0,1/2]$, both $x$ and $x+z$
belong to $(-\infty,0]$ and convexity implies $U(x+z)\leq U(x)$. 
Thus $\pi(x+z)\geq\pi(x)$ and
\[
\alpha(x)
\geq
\int_0^{1/2}\phi_c(z)\,dz
\geq
\frac12\phi_c(1/2).
\]
Similarly for $x>1/2$ by taking $z\in[-1/2,0]$.
    
\noindent\textit{Case 3: $y=U^\leftarrow$ and $x\in(-1/2,U^\leftarrow)$.}
For every $w\in[0,1/2]$, we have $U(w)\leq U(1/2)< U(x)$ by definition of $U^\leftarrow$, so $\pi(w)\geq\pi(x)$. Therefore,
\begin{align*}
\alpha(x)
&\ge \int_{-x}^{-x+1/2} \min\left\{
1,\frac{\pi(x+z)}{\pi(x)}
\right\}
\phi_c(z)\,dz
\\
&= \int_{0}^{1/2} \min\left\{
1,\frac{\pi(w)}{\pi(x)}
\right\}
\phi_c(w-x) \,dw\\
&= \int_{0}^{1/2} \phi_c(w-x) \,dw \geq
\frac12\phi_c(1).
\end{align*}
Combining the three cases gives
\[
\alpha(x)
\geq
\min\left\{
\frac{\phi_c(1/2)}{2},
\frac{\phi_c(1)}{2},
\frac{e^{-2.6}\phi_c(3/2)}{2}\right\}.
\] 
Since $\phi_c$ is nonincreasing on $[0,\infty)$ and $e^{-2.6}\leq1$,
the minimum is attained by the third term. Therefore,
\[
\alpha(x)\geq
\phi_c(3/2)\frac{e^{-2.6}}{2}
=b(c).
\]
\end{proof}

\begin{lemma}\label{boundU1}
Let $\pi$ be as in Assumption \ref{assA} with $ U(0) = U'(0) = 0 $, and let $s^2=\var(\pi)$ denote the variance. Then 
\[\min\{U(-s),U(s)\}\leq 2.6.\]
\end{lemma}
\begin{proof}
Without loss of generality, we may assume that $s^2 = 1$. If this is not the case, we can define a rescaled density \(\tilde{\pi}(x) = s \pi(s x)\), which has variance \(\operatorname{Var}(\tilde{\pi}) = 1\). The associated potential becomes \(\tilde{U}(x) = U(s x)\), and thus
\[
\min\{U(-s), U(s)\}=\min\{\tilde{U}(-1), \tilde{U}(1)\}  \leq 2.6.
\]
The first step of the proof is to bound the variance. Since $\var(\pi)=s^2=1$, we have
    \[
    1\leq 2\max\left\{\int_{-\infty}^0 x^2\pi(x)dx,\int_0^{+\infty}x^2\pi(x)dx\right\}.
    \]
    Without loss of generality, we assume that the maximum is attained on the positive half-line. We can decompose the integral into the sum of two terms:
    \[
    \int_0^{+\infty}x^2\pi(x)dx=\int_0^1x^2\pi(x)dx+\int_1^{+\infty}x^2\pi(x)dx.
    \]
    We bound the first term as 
    \begin{equation}\label{bound1}
        \int_0^1x^2\pi(x)dx\stackrel{i)}\leq\int_0^1 x^2\frac{e^{-U(x)}}{\int_0^1e^{-U(x)}dx}dx\stackrel{ii)}\leq \int_0^1x^2dx=\frac{1}{3},
    \end{equation}
    where $i)$ comes from the fact that the normalizing constant $N:=\int_\R e^{-U(x)}dx$ is lower bounded by $N\geq\int_0^1e^{-U(x)}dx$, and for $ii)$ we apply Lemma \ref{stoch_ord}, comparing the probability density functions $f(x)=\frac{e^{-U(x)}}{\int_0^1e^{-U(x)}dx}$ and $g(x)=1$ on $[0,1]$, with the non-decreasing function $\bar{\eta}(x)=x^2$.\\
    For the second term, we begin by bounding $N$. For $x\in(0,1)$, convexity of $U$ implies $U(x)\leq U(1)x$, so
    \[
    N\geq\int_0^1e^{-U(x)}dx\geq\int_0^1 e^{-U(1)x}dx=\frac{1-e^{-U(1)}}{U(1)}.
    \]
    Furthermore, the convexity of $U$ implies that $U'$ is increasing, so $U(1)=\int_0^1U'(x)dx\leq U'(1)$. Using convexity again, for $x\geq1$ we obtain 
    \[
    U(x)\geq U(1)+U'(1)(x-1)\geq U(1)x.
    \]
    Hence, we can estimate:
    \begin{align*}\label{bound0}
    \int_1^{+\infty}x^2\pi(x)dx&\leq \frac{U(1)}{1-e^{-U(1)}}\int_1^{+\infty}x^2e^{-U(1)x}dx\\
    &=\frac{U(1)}{1-e^{-U(1)}}\frac{e^{-U(1)}}{U(1)^3}\left(U(1)^2+2U(1)+2\right)\\
    &\leq\frac{1}{e^{U(1)}-1}\left(1+\frac{2}{U(1)}+\frac{2}{U(1)^2}\right).
    \end{align*}
    We define the function $h:\R^+\rightarrow\R^+$, $h(x)=\frac{1}{e^x-1}\left(1+\frac{2}{x}+\frac{2}{x^2}\right)$, so we rewrite the last inequality as 
    \begin{equation}\label{bound2}
        \int_1^{+\infty}x^2\pi(x)dx\leq h(U(1)).
    \end{equation}
    Combining inequalities \eqref{bound1} and \eqref{bound2}, we obtain $1\leq 2 \left(\frac{1}{3}+h(U(1))\right)$, that implies $h(U(1))\geq 1/6$.
    The function $h$ is continuous, strictly decreasing on $(0,\infty)$, with $\lim_{x\rightarrow 0^+}h(x)=+\infty$ and $\lim_{x\rightarrow +\infty}h(x)=0$, so it is bijective and it implies that
    \[U(1)\leq h^{-1}(1/6)<2.6.\]
    This completes the proof.
\end{proof}

\begin{proof}[Proof of Theorem \ref{result1}(b)]
    In the one-dimensional case, $\E[\|X-\E [X]\|^2]$ is equal to $\var(\pi)$. The proof is completed by substituting this into Theorem \ref{result1}(a), and using the fact that $\sigma=\sqrt{c\var(\pi)}$, along with the lower bound on $\alpha_0$ from Proposition \ref{boundacceptance}.
\end{proof}

\begin{lemma}\label{stoch_ord}
Let $f$ and $g$ be two probability densities on $\R$ such that $g(x)/f(x)$ is non-decreasing in $x$.
Then $\int \bar{\eta}(x)f(x)\mathrm{d}x\leq \int \bar{\eta}(x)g(x)\mathrm{d}x$ for every non-decreasing $\bar{\eta}:\R\to\R$.
\end{lemma}
Lemma \ref{stoch_ord} is a well-known result of stochastic ordering. A proof can be found for example in \citet[Lemma S.1]{ascolani2023clustering}.
 
\section{Metropolis-within-Gibbs}\label{section3}
We now apply the results of the previous section to MwG schemes, which are our motivating application.

\subsection{Notation and definitions}
First, we provide the notation required to define the Markov transition kernel of MwG.
For each $x=(x_1,\dots,x_d)\in\R^d$ and $m\in\{1,\dots,d\}$, we denote the vector obtained by removing the $m$-th component as 
\[
x_{-m}=(x_1,\dots,x_{m-1},x_{m+1},\dots,x_d)\in\R^{d-1}.
\]
Given $\mu\in\mathcal{P}(\R^d)$ and $X\sim\mu$, we denote by $\mu(\cdot|x_{-m})\in\mathcal{P}(\R)$ the conditional distribution of $X_m$ given $X_{-m}=x_{-m}$.

The Markov transition kernel associated to a random-scan MwG algorithm takes the form
\begin{equation}\label{eq:mwg}
    P^\text{MwG}=\frac{1}{d}\sum_{m=1}^dP_m,\quad\quad P_m(x,dy)=P_m^{x_{-m}}(x_m,dy_m)\delta_{x_{-m}}(dy_{-m})\,,
\end{equation}
where $P_m^{x_{-m}}$ is an arbitrary $\pi(\cdot|x_{-m})$-invariant Markov kernel on $\R\times\mathcal{B}(\R)$. 
Thus, at each iteration, the Markov chain governed by $P^\text{{MwG}}$ selects a coordinate $m\in\{1,\dots,d\}$ uniformly at random and updates the $m$-th component of $x$ via the kernel $P_m^{x_{-m}}$, leaving the other coordinates unchanged. 

Typically, each conditional kernel $P_m^{x_{-m}}(x_m,dy_m)$ is implemented via a MH step, with proposal distribution $Q_m^{x_{-m}}(x_m,\cdot)\in\mathcal{P}(\R)$ and transition kernel
\begin{equation}\label{MwGsampler}
P_m^{x_{-m}}(x_m,dy_m)=\alpha_m^{x_{-m}}(x_m,y_m)Q_m^{x_{-m}}(x_m,dy_m)+r^{x_{-m}}_m(x_m)\delta_{x_m}(dy_m),
\end{equation}
where
\[
\alpha_m^{x_{-m}}(x_m,y_m)=\min\left\{1,\frac{\pi(dy_m|x_{-m})Q_m^{x_{-m}}(y_m,dx_m)}{\pi(dx_m|x_{-m})Q_m^{x_{-m}}(x_m,dy_m)}\right\}
\]
and 
$r_m^{x_{-m}}(x_m)=1-\int\alpha_m^{x_{-m}}(x_m,y_m)Q_m^{x_{-m}}(x_m,dy_m)$.

\begin{proposition}\label{prop:mwg-positive}
Let $P^\text{MwG}$ be a random-scan MwG kernel as in \eqref{eq:mwg}, such that for every $m$ and for $\pi_{-m}$-almost every $x_{-m}$, the conditional kernel $P_m^{x_{-m}}$ is positive on $L^2(\pi(\cdot\mid x_{-m}))$. Then $P^\text{MwG}$ is positive on $L^2(\pi)$. 
In particular, this holds for the RWM conditional kernels considered in Theorem \ref{thmprincipalresult}.
\end{proposition}

\begin{proof} 
For $f\in L^2(\pi)$, write $f_{x_{-m}}(x_m):=f(x_m,x_{-m})$. 
Using the disintegration of $\pi$, we have \begin{align*} \langle f,P_m f\rangle_\pi = \int_{\mathbb{R}^{d-1}} \left\langle f_{x_{-m}}, P_m^{x_{-m}}f_{x_{-m}} \right\rangle_{\pi(\cdot\mid x_{-m})} \,\pi_{-m}(dx_{-m})\geq 0. \end{align*} 
We conclude that $P^{\mathrm{MwG}}$ is positive as a weighted sum of positive kernels.

For the RWM case, by Proposition \ref{prop:rwm-positive}, for every $m,x_{-m}$, the kernel $P_m^{x_{-m}}$ is positive on $L^2(\pi(\cdot\mid x_{-m}))$.
\end{proof}

\subsection{Spectral gap lower bound for MwG with RWM updates}
We can now derive a sharper lower bound on the spectral gap of the MwG Markov kernel when the updates are performed via a RWM step. Specifically, we improve the condition number dependence of the best previously available bound \citep[Corollary 7.4]{ascolani2024entropy}, which shows that $\gap(P^\text{MwG})=\Omega((\kappa^2 d)^{-1})$, by proving a stronger bound of the form $\gap(P^\text{MwG})=\Omega((\kappa d)^{-1})$, where $g=\Omega(f)$ means $g\geq c f$ for a constant $c>0$.

We consider the following class of target distributions, which originally comes from the literature on the analysis of coordinate descent algorithms for convex optimization \citep[Sec. 3]{nesterov2012efficiency} which allow the convergence bound to be expressed through a coordinate-dependent condition number.
\begin{assumption}\label{assB}
$\pi\in\mathcal{P}(\R^d)$ has density $\pi(x)\propto e^{-U(x)}$, where $U:\R^d\rightarrow\R$ is of class $\mathcal{C}^1$ and satisfies:
\begin{itemize}
    \item for every $m=1,\dots,d$ and $x_{-m}\in\R^{d-1}$, the function $x_m\mapsto\nabla_m U(x_m,x_{-m})$ is $L_m$-lipschitz;
    \item the function $x\mapsto U(x)-\frac{\lambda^*}{2}\|x\|_L^2$ is convex, where $\|x\|_L^2:=\sum_m L_m|x_m|^2$ and $\lambda^* >0$.
\end{itemize}
We define $\kappa^*:=1/\lambda^*$.
\end{assumption}

This condition is less restrictive than the classical strong log-concavity and smoothness hypothesis on $\pi$; specifically, if $\pi$ has a potential $U$ that is $L$-smooth and $m$-strongly convex, with condition number $\kappa = L/m$, then it also satisfies Assumption \ref{assB} with
\[
1 \leq \kappa^* \leq \kappa,
\]
see e.g.\ \citet[Lemma 2.4]{ascolani2024entropy}.
Consequently, Theorem \ref{thmprincipalresult} remains valid even if we assume strong log-concavity and smoothness and replace $\kappa^*$ with $\kappa$. However, the version with $\kappa^*$ can lead to tighter convergence guarantees in specific applications (see e.g.\ \citet{ascolani2025mixing} for applications to data augmentation samplers).
The underlying intuition is that Assumption \ref{assB} is tailored to the coordinate-wise setting, and can thus be more refined for analyzing algorithms that operate through coordinate-wise updates.

\begin{theorem}[MwG with RWM updates]\label{thmprincipalresult}
Let $\pi$ satisfy Assumption \ref{assB}.
For every $m=1,\dots,d$ and $x_{-m}\in\mathbb{R}^{d-1}$, let $P_m^{x_{-m}}$ be the RWM kernel targeting $\pi(\cdot\mid x_{-m})$ as in \eqref{MwGsampler} with 
\[
Q_m^{x_{-m}}(x_m,\cdot) = \mathcal{N}\bigl(x_m,\sigma_m^2(x_{-m})\bigr), \qquad \sigma_m^2(x_{-m}) = c_m(x_{-m}) \operatorname{Var}\bigl(\pi(\cdot\mid x_{-m})\bigr). 
\]
Suppose that there exist constants $0<\underline{c}\leq \overline{c}<\infty$ such that $\underline{c} \leq c_m(x_{-m}) \leq\overline{c}$ for every $m=1,\dots,d$ and $x_{-m}\in\mathbb{R}^{d-1}$.

Then the spectral gap of the corresponding random-scan MwG kernel satisfies 
\begin{equation}\label{eq:lower-bound}
\operatorname{Gap}\bigl(P^{\operatorname{MwG}}\bigr) \geq \frac{C_{\underline{c},\overline{c}}}{\kappa^*d},     
\end{equation}
where $C_{\underline{c},\overline{c}} := \frac{1}{2} \min_{c\in[\underline{c},\overline{c}]} k(c)^2 >0$, and $k(\cdot)$ is defined in Theorem \ref{result1}(b).
\end{theorem}
\begin{proof}
MwG is reversible by construction and positive on $L^2(\pi)$ by Proposition \ref{prop:mwg-positive}.
By Assumption \ref{assB}, for every $m\in\{1,\dots,d\}$ and $x_{-m}\in\R^{d-1}$, the conditional distribution $\pi(\cdot|x_{-m})$ is log-concave. 
Thus, by \eqref{lowerboundspectral}, $P_m^{x_{-m}}$ satisfies
\begin{equation}\label{eq:proof-pt1}
    \gap(P_m^{x_{-m}})\geq \frac{k(c_m(x_{-m}))^2}{2},
\end{equation}
with $k(\cdot)$ defined in Theorem \ref{result1}(b).
By \citet[Theorem 7.1(b)]{ascolani2024entropy}, the spectral gap of the overall MwG kernel satisfies
\begin{equation}\label{relationgaps}
\gap(P^\text{MwG})\geq \inf_{m,x_{-m}}\gap(P_m^{x_{-m}})\cdot\gap(P^\text{GS}),    
\end{equation}
and by \citet[Corollary 3.7]{ascolani2024entropy}, 
\[
\gap(P^\text{GS})\ge\frac{1}{\kappa^* d}.
\]
Combining it with \eqref{eq:proof-pt1} yields \eqref{eq:lower-bound}.
\end{proof}

\begin{remark}[Tuning of proposal variances]\label{rmk:uniform-tuning}
Theorem \ref{thmprincipalresult} assumes that the proposal variances are uniformly comparable to the corresponding conditional variances. More precisely, 
\[ 
\underline{c}\, \var\bigl(\pi(\cdot\mid x_{-m})\bigr) \leq \sigma_m^2(x_{-m}) \leq \overline{c}\, \var\bigl(\pi(\cdot\mid x_{-m})\bigr) 
\]
for every $m=1,\dots,d$ and $x_{-m}\in\mathbb{R}^{d-1}$, where $0<\underline{c}\leq\overline{c}<\infty$. The constants $\underline{c}$ and $\overline{c}$ quantify the uniform calibration of the proposal step sizes and determine the constant $C_{\underline{c},\overline{c}}$ appearing in the lower bound \eqref{eq:lower-bound}.
By contrast, if the scaling coefficients are allowed to approach either $0$ or $+\infty$, the resulting constant may deteriorate to zero.
\end{remark}

\begin{remark}[Tightness of the bound]\label{rmk:tightness}
For MwG with RWM updates it holds
    \[
    \gap(P^\text{MwG})\leq \gap(P^\text{GS})\,,
    \]
see e.g.\ \citep[Corollary 4]{qin2025spectral}. Also, when $\pi$ is multivariate normal, $\gap(P^\text{GS})=(\kappa^*d)^{-1}$ \citep[Theorem 1]{amit1996convergence}. Combining these results implies that, at least for the family of multivariate normal distributions, the bound in Theorem \ref{thmprincipalresult} is tight up to up to constants depending only on the uniform tuning bounds
$\underline c$ and $\overline c$.
\end{remark}

\subsection{Mixing times}
It is well-known that the spectral gap of a positive, $\pi$-reversible Markov kernel $P$ directly controls the convergence rate in $\chi^2$-divergence. Specifically, for any initial distribution $\mu^{(0)}\ll\pi$ and any $n\in\mathbb{N}_0$, the iterates $\mu^{(n+1)}=\mu^{(n)}P$, with
\[
\mu^{(n)}P(A)=\int P(x,A)\mu^{(n)}(dx)\quad \text{for all }A\in\mathcal{B}(\R^d),
\]
satisfy the following relation, that can be deduced from \eqref{conv},
\begin{equation}\label{conv2}
d_2(\mu^{(n)},\pi)\leq (1-\gap(P))^{n}\,d_2(\mu^{(0)},\pi),    
\end{equation}
where $d_2$ is the square-root of the $\chi^2$-divergence,
\[
d_2(\mu,\pi)=\left(\int\left(\frac{d\mu}{d\pi}-1\right)^2 d\pi\right)^{1/2}.
\] 

By Proposition \ref{prop:mwg-positive} the MwG kernel with RWM updates $P^{\mathrm{MwG}}$ is
positive, so \eqref{conv} and hence \eqref{conv2} apply to it.
Thus, to ensure $d_2(\mu^{(n)},\pi)\leq \varepsilon$, by Theorem \ref{thmprincipalresult} it suffices to take
\[
n\geq \frac{\kappa^* d}{C_{\underline{c},\overline{c}}}\log\left(\frac{d_2(\mu^{(0)},\pi)}{\varepsilon}\right).
\]
The above bound depends on the logarithm of the initial distance $d_2(\mu^{(0)},\pi)$. 
In the case of an $\eta$-warm start initialization, that is, when there exists $\eta\geq1$ such that 
\[
\sup_{A\in\mathcal{B}(\R^d),\pi(A)\neq 0}\frac{ \mu^{(0)}(A)}{\pi(A)}\leq\eta,
\]
the resulting mixing time is bounded by
\[
n\geq\frac{\kappa^* d}{C_{\underline{c},\overline{c}}}\log\left(\frac{\eta}{\varepsilon}\right).
\]
This result is sub-optimal in terms of dependence on $\eta$, especially when $\eta$ grows exponentially with the dimension $d$. This situation arises, for example, in the commonly studied case of a feasible start when $\pi$ is log-concave (see e.g. \citet[Equation~(12)]{dwivedi2019log}). 
Improved dependence on $\eta$, namely a double-logarithmic one, has been obtained for different Markov chains through a stronger isoperimetric inequality for $\pi$, combined with a more refined conductance analysis. This approach makes use of the \textit{conductance profile}, which extends the classical notion of conductance by tracking its behavior across families of subsets with the same measure, see \citet[Section~3.3]{chen2020fast}. The technique is by now well established and has been used, for example, in \cite{andrieu2024explicit} and \cite{lee2024fast}. Obtaining a similar dependence on $\eta$ for MwG is an interesting direction for 
future work that would potentially require some non-trivial extension of the comparison bounds in \cite{qin2025spectral,ascolani2026scalability,qin2024spectral} to conductance profiles.

\section{An application to hierarchical logistic regression models}\label{sec:application}
We illustrate the improvement in spectral gap bounds achieved by our method through a concrete example. 
Specifically, we consider a classical Bayesian hierarchical model \citep{gelman1995bayesian} where the observed data $Y$ are grouped into $J$ distinct groups, each associated with a latent parameter $\theta_j$. Within each group $j$, we observe $n$ binary outcomes $Y_{j1},\dots,Y_{jn}$, modeled as conditionally independent given $\theta_j$:
\begin{align}
    Y_{ji}\,|\,\theta_j &\stackrel{iid}\sim \operatorname{Bernoulli}\left(\frac{e^{\theta_j}}{1+e^{\theta_j}}\right) \quad&&j=1,\dots,J,\quad i=1,\dots,n \notag \\
    \theta_j\,|\,\mu &\stackrel{iid}\sim \mathcal{N}(\mu,1) && j=1,\dots,J \label{modellog}\\
    \mu&\sim \mathcal{N}(0,1).   \notag
\end{align}
The target distribution of the algorithm is the full posterior $\pi^y(\theta,\mu)$, defined as the conditional distribution of $(\theta,\mu)$ given $Y=y$ under \eqref{modellog}. 
By the hierarchical structure, the posterior distribution of $\theta$ conditional on $\mu$ factorizes as $\pi^y(\theta\,|\,\mu)=\prod_{j=1}^J\pi^y(\theta_j\,|\,\mu)$, so that the variables $\theta_j$ are conditionally independent given $\mu$. Consequently, the update of $\theta_j$ can be performed independently of $\theta_{-j}$, and its target has the expression
\begin{align*}
\pi^y(\theta_j\,&|\,\theta_{-j},\mu)\propto
p(y_{j1},\dots,y_{jn}\,|\,\theta_j)\,p(\theta_j\,|\,\mu)\\
    &\propto \exp\left(y_j\theta_j-n\log(1+e^{\theta_j})-\frac{(\theta_j-\mu)^2}{2}\right)=: \exp\left(-U_j(\theta_j)+\text{const}\right),
\end{align*}
where $y_j=\sum_{i=1}^n y_{ji}$ and $U_{j}$ is the associated potential.
Recalling that the condition number $\kappa$ of a twice-differentiable potential $U:\R\longrightarrow \R$ is defined as
\[
\kappa(U)=\frac{\sup_x U''(x)}{\inf_x U''(x)}=\frac{L}{m},
\]
we can explicitly compute the condition number associated with $\pi^y(\theta_j\,|\,\theta_{-j}, \mu)$ (see \ref{compkappa}), obtaining
\begin{equation}\label{eq:cond_num_log}
\kappa(U_{j})=1+\frac{n}{4}.    
\end{equation}
For $\mu$, we have that 
\begin{equation}\label{postgauss}
\pi^y(\mu\,|\,\theta)\sim\mathcal{N}\left(\frac{\sum_{j=1}^J\theta_j}{J+1},\frac{1}{J+1}\right),    
\end{equation}
thus 
\[U_0(\mu):=-\log(\pi^y(\mu\,|\,\theta))=\frac{J+1}{2}\mu^2-\left(\sum_{j=1}^J\theta_j\right)\mu+\text{const},
\]
so that 
$U''_0(\mu)= J+1$ and its condition number is equal to $\kappa(U_0)=1$.

We now analyze the MwG with RWM updates with target distribution $\pi^y(x)$, where
$x=(\mu,\theta_1,\dots,\theta_J)\in\R^{J+1}$, under two different proposal kernels. This allows us to compare the bound previously available with the one introduced in this paper.

In the first scenario, we consider a common choice in theoretical analyses of MCMC, where the variance of the proposal kernel is chosen to be proportional to the inverse of the smoothness parameter, so that the 1-dimensional proposal is
\[Q_j^{x_{-j}}(x_j,\cdot)=\mathcal{N}(x_{j},c_j/L_j),\]
where $c_j>0$ is a constant.
In this case, the best available bound, namely \citet[Corollary 35]{andrieu2024explicit}, implies $\gap(P_j^{x_{-j}})\geq \frac{\tilde{C}}{\kappa(U_j)}$, where $\tilde{C}$ is a universal constant and $P_j^{x_{-j}}$ is the MH kernel with proposal $Q_j^{x_{-j}}$ and target $\pi^y(x_j|x_{-j})$.
Combined with the spectral gap decomposition \eqref{spectralgapdec} and with \eqref{eq:cond_num_log}, 
this leads to the lower bound
\begin{equation*}
\gap(P^\text{MwG})\geq \min_j \frac{\tilde{C}}{\kappa(U_j)}\cdot\gap(P^\text{GS})=\frac{\tilde{C}}{1+n/4}\cdot\gap(P^\text{GS}).    
\end{equation*}
This bound deteriorates as the number $n$ of data points per parameter increases, suggesting that MwG should become progressively slower relative to GS as $n$ grows. 
However, this is inconsistent with the empirical evidence (see e.g.\ Figure \ref{fig:iat-boxplots} below). 

In the second scenario, we use the proposal 
\[Q_j^{x_{-j}}(x_j,\cdot)=\mathcal{N}(x_j,\sigma^2_j(x_{-j}))\]
with $\sigma^2_j(x_{-j})=c_j(x_{-j})\var(\pi(\cdot|x_{-j}))$ and $c_j(x_{-j})\in[\,\underline{c},\overline{c}\,]$. Theorem \ref{thmprincipalresult} yields 
\[
\gap(P^\text{MwG})\geq C_{\underline{c},\overline{c}}\,\gap(P^\text{GS}),
\]
where $C_{\underline{c},\overline{c}}$ depends only on the uniform tuning bounds $\underline c$ and $\overline c$, and in particular is independent of both $n$ and $J$.
This bound is larger by a factor $n$ than the previous one and provides theoretical support to the empirical observations that MwG incurs only a small slowdown relative to GS for this model, even as $n$ grows.

\subsection{Numerical simulations}

We consider the hierarchical logistic model \eqref{modellog} and investigate how the mixing performance of MwG samplers with RWM updates changes as the number of observations per group increases. We consider $J\in\{10,100\}$ groups and
$n\in\{2^5,\ldots,2^9\}$ observations per group.

To measure mixing performance, we use the integrated autocorrelation time (IAT).
For a $\pi$-invariant Markov chain $(X_0,X_1,\dots)$ and a square-integrable function $f\in L^2(\pi)$, the IAT associated with $f$ is defined as
\[
\text{IAT}(f):=1+2\sum_{t=1}^\infty\text{Corr}\left(f(X_0),f(X_t)\right).
\] 
This quantity controls the asymptotic variance of the MCMC sample mean estimator $\bar{f}=\frac{1}{m}\sum_{t=1}^mf(X_t)$ through the relation
\[
\var(\bar{f})= \text{IAT}(f)\frac{\var_\pi(f)}{m}+o\left(\frac{1}{m}\right)\,,
\]
provided that the correlations $\text{Corr}(f(X_0),f(X_t))$ decay exponentially fast as $t\rightarrow\infty$ \citep{sokal1997monte}.
Thus, $\text{IAT}(f)$ can be interpreted as the number of correlated MCMC samples required, once the chain has reached stationarity, to match the information content of one independent draw from the target distribution $\pi$.
We adopt the IAT as our measure of mixing performance because of its direct connection with the spectral gap: specifically, $2/\gap(P)$ provides an upper bound on the worst case IAT, see e.g.\ \citep[Proposition 1]{rosenthal2003asymptotic}, that is tight up to a factor of $2$ \citep{kipnis1986central}.

\paragraph{Sampling schemes}
In all the algorithms considered below, the global parameter $\mu$ is sampled exactly from its Gaussian full conditional distribution \eqref{postgauss}. For the group-specific parameters $\theta_j$, we compare the following three MwG schemes.

In the first method, for each coordinate, the proposal variance is set to $\sigma_j^2(\mu)=10   \,\var(\theta_j\mid y_j,\mu)$; the conditional variance is recomputed at the current value of $\mu$ using Gauss-Hermite quadrature, as implemented in the R package \texttt{statmod}. 
We verified numerically that $5$ quadrature nodes approximate the conditional variance with a relative error of order $10^{-3}$ with respect to the reference value, obtained with $41$ nodes; we regarded this accuracy as sufficient for the present experiment.
In the second method, the proposal variance is adapted during burn-in using a Robbins-Monro stochastic approximation scheme targeting an acceptance probability of \(0.4\); see \citet{andrieu2008tutorial}. Adaptation is stopped after burn-in, and the resulting proposal variances are held fixed during the sampling phase. In the third method, the proposal variance is set to $\sigma_j^2(\mu)=17.5/L$, where $L=(n+4)/4$ is the smoothness constant of the coordinate potential in \eqref{maxU} and the constant $17.5$ has been chosen after a pilot run. As a benchmark, we also consider the GS, in which draws from the log-concave full conditional distributions of the $\theta_j$'s are generated by adaptive rejection sampling (ARS) \citep{gilks1992adaptive}, as implemented in the R package \texttt{ars}.

\paragraph{Simulation design} 
For every combination of $J$ and $n$, we generate $100$ independent datasets under the model with $\mu^\star=1$, and within each replication, the same simulated dataset is used for all four sampling schemes.
Each chain is run for $1000$ burn-in iterations, followed by $4000$ sampling iterations. 
We estimate the IAT as the ratio between the number of iterations and the effective sample size (ESS) \citep{gong2014practical}, computed using the \texttt{ess} function from the R package \texttt{mcmcse}. For each replication, we then record the maximum coordinate-wise IAT.

\begin{figure}[t]
  \centering
  \begin{subfigure}[t]{0.49\textwidth}
    \centering
    \includegraphics[width=\linewidth]{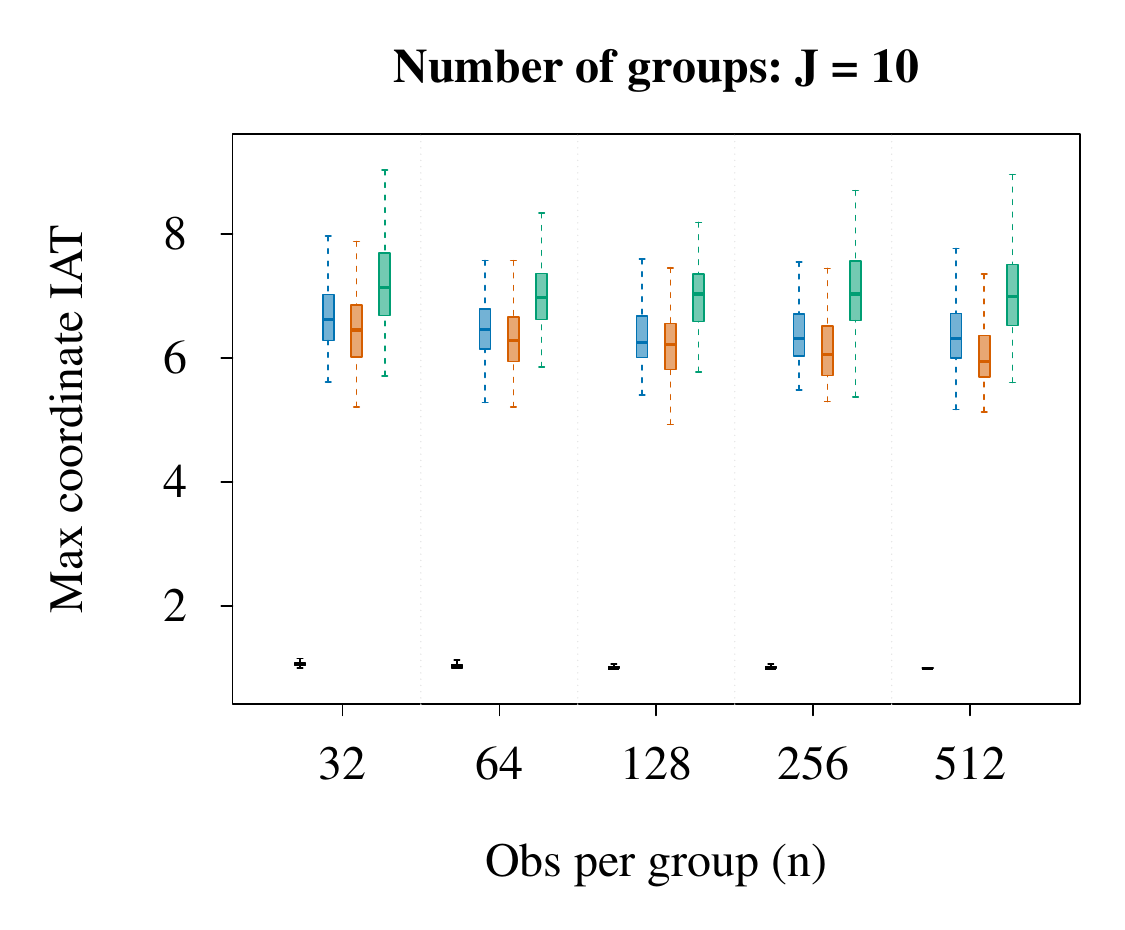}
  \end{subfigure}\hfill
  \begin{subfigure}[t]{0.49\textwidth}
    \centering
    \includegraphics[width=\linewidth]{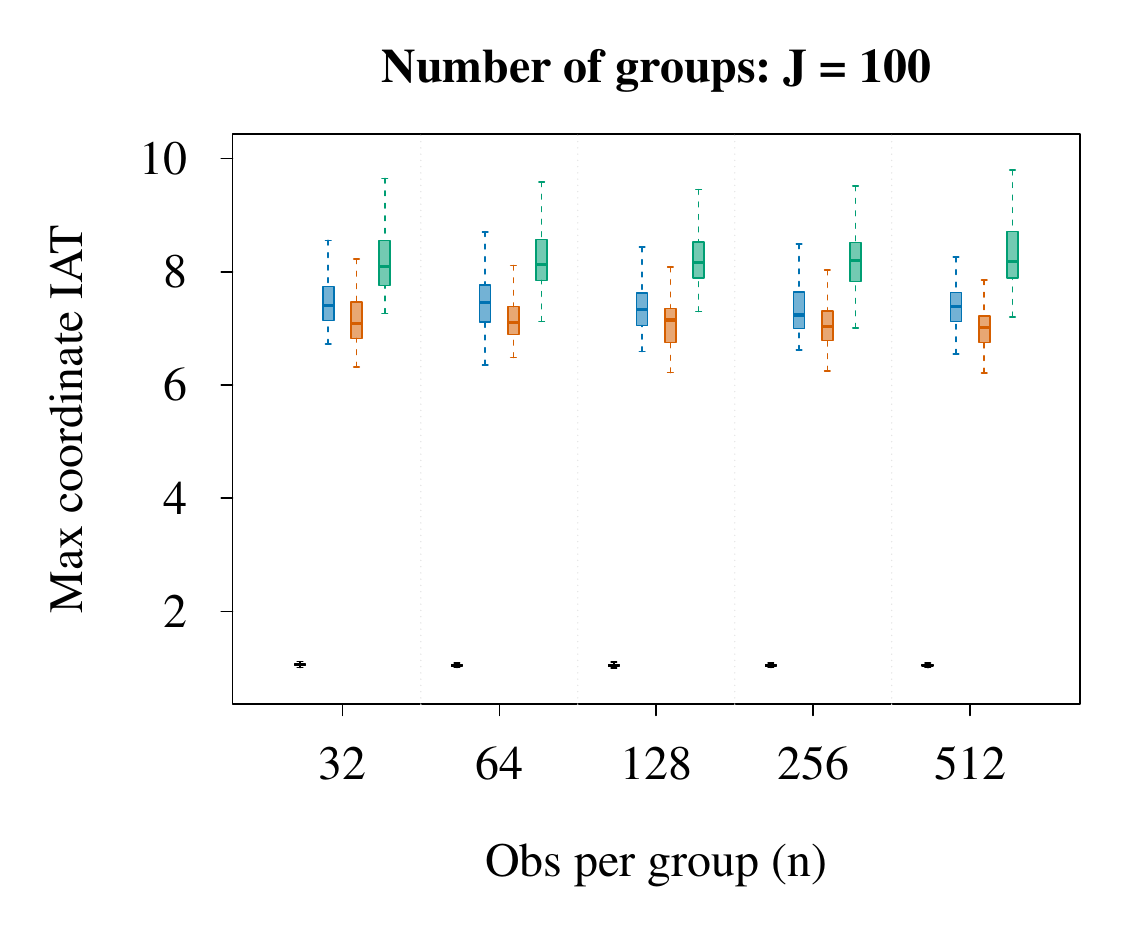}
  \end{subfigure}
  
  \vspace{-3em}
  
  \includegraphics[width=0.9\textwidth]{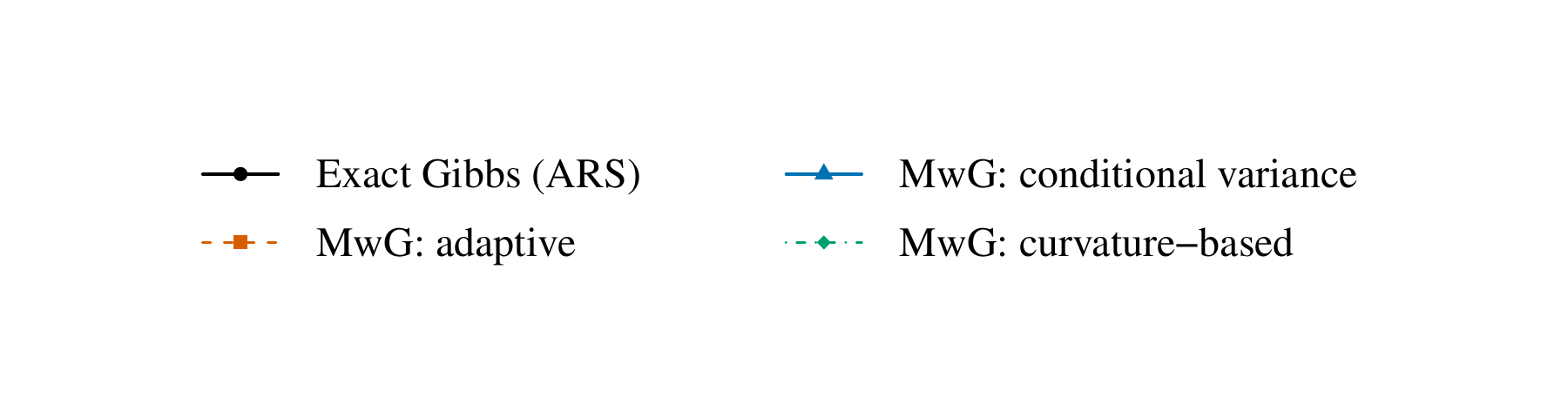}
  
  \vspace{-2em}
\caption{Boxplots, across $100$ independently generated datasets, of the maximum estimated coordinate-wise IAT for the GS and the three MwG schemes, plotted against the number $n$ of observations per group. The left and right panels correspond to $J=10$ and $J=100$, respectively.}
  \label{fig:iat-boxplots}
\end{figure}

\paragraph{Results}
Figure \ref{fig:iat-boxplots} shows that for both $J=10$ and $J=100$ the maximum coordinate-wise IAT of each MwG scheme remains essentially stable over the considered range of $n$. In particular, although the MwG samplers have larger IATs than the GS, the difference remains approximately constant as $n$ increases. 
The additional plots in \ref{app:extra-plots} show the average proposal standard deviations and acceptance rates.

These findings are consistent with the theoretical comparison in the previous section. In this example, conditional-variance scaling yields a lower bound on the ratio between the MwG and GS spectral gaps whose constant does not deteriorate with $n$; correspondingly, the simulations show an approximately stable MwG-to-GS slowdown over the considered range of $n$. 
By contrast, the previously available comparison introduces an additional factor of order $n$ in the upper bound on the relaxation time of MwG relative to that of GS, and is therefore overly conservative.

The conditional-variance scheme provides the most direct numerical illustration of the theoretical tuning rule, but it requires repeatedly approximating the conditional variances and is consequently more computationally demanding. The adaptive and curvature-based schemes avoid this repeated calculation while achieving comparable IATs. This suggests that they offer practically useful approximations to conditional-variance tuning in this example.

\appendix
\section{Computation of \texorpdfstring{$\kappa$}{k} for \texorpdfstring{$\pi^y(\theta_j|\theta_{-j},\mu)$}{pt}}\label{compkappa}
The potential of the conditional distribution $\pi^y(\theta_j|\theta_{-j},\mu)$ is given by
\[U(\theta_j)=-y_j\theta_j+n\log(1+e^{\theta_j})+\frac{(\theta_j-\mu)^2}{2},\]
with second derivative
\begin{align*}
U''(\theta_j)&=\frac{n e^{\theta_j}}{(1+e^{\theta_j})^2}+1.    
\end{align*}
To compute the condition number of $\pi^y(\theta_j|\theta_{-j},\mu)$, it suffices to study the supremum and infimum of $U''(\theta_j)$.
The third derivative of $U$ is 
\[
U'''(\theta_j)=-\frac{ne^{\theta_j}(e^{\theta_j}-1)}{(e^{\theta_j}+1)^3},
\]
which shows that $U''$ is increasing for $\theta_j<0$, attains its maximum at $\theta_j=0$, and decreases for $\theta_j>0$. Consequently the maximum is equal to
\begin{equation}\label{maxU}
    \sup_{\theta_j}U''(\theta_j)=U''(0)=1+\frac{n}{4}
\end{equation}
and the infimum is approached for $|\theta_j|\rightarrow+\infty$; specifically 
\[
\lim_{\theta_j\rightarrow\pm\infty}U''(\theta_j)=1.
\]
Therefore, the condition number is 
\[
\kappa(U_{\theta_j})=\frac{\sup_\theta U''(\theta)}{\inf_\theta U''(\theta)}=\frac{n/4+1}{1}=\frac{n}{4}+1.
\]

\section{Extra plots about the simulations}\label{app:extra-plots}

\begin{figure}[t]
  \centering
  \begin{minipage}[c]{\textwidth}
    \centering
    \begin{subfigure}[t]{0.50\linewidth}
      \centering
      \includegraphics[width=\linewidth]{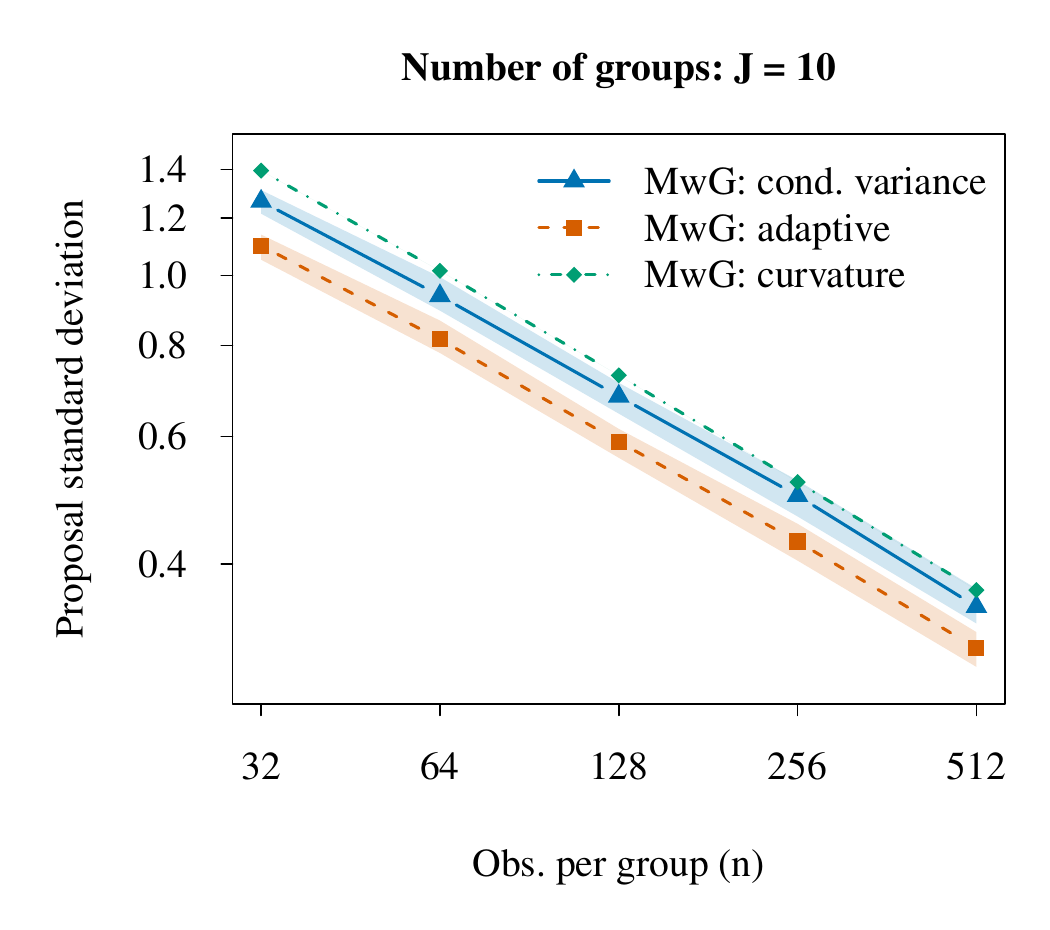}
    \end{subfigure}\hfill
    \begin{subfigure}[t]{0.50\linewidth}
      \centering
      \includegraphics[width=\linewidth]{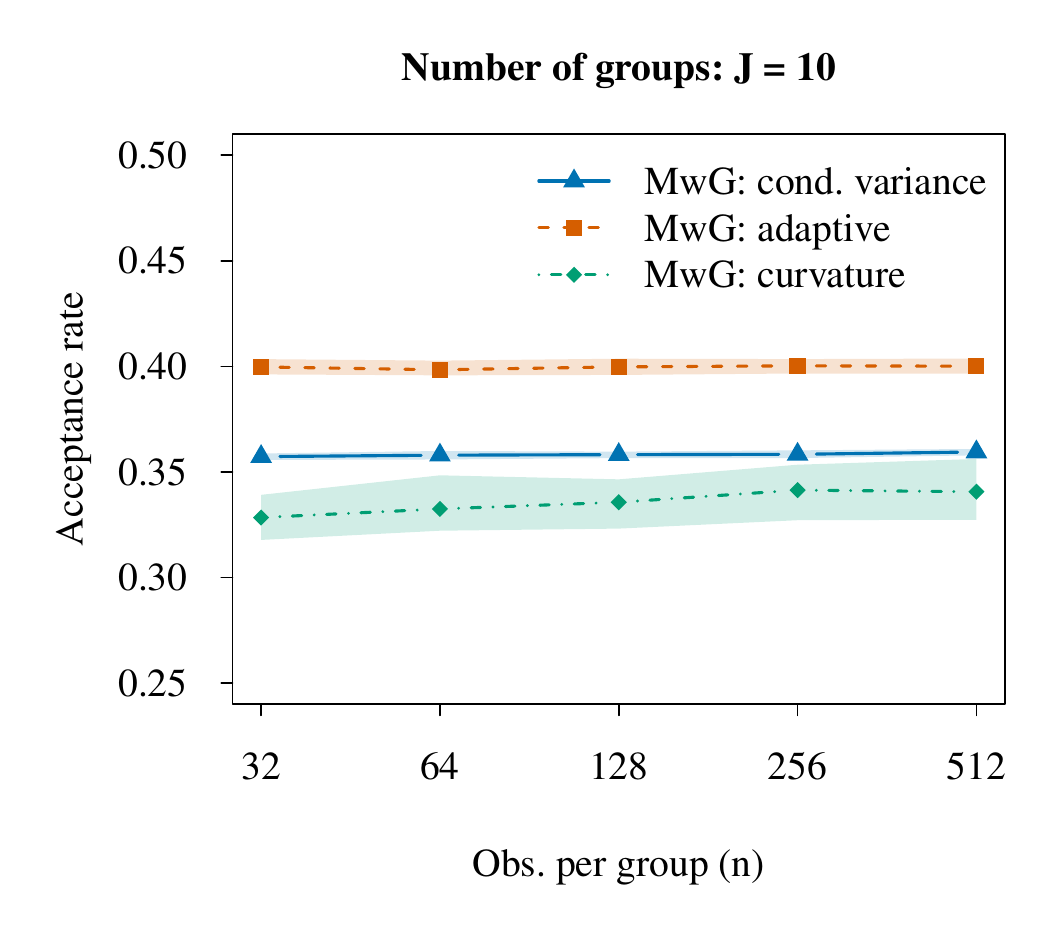}
    \end{subfigure}
  \end{minipage}\hfill
  \centering 
  \begin{minipage}[c]{\textwidth}
    \centering
    \begin{subfigure}[t]{0.5\linewidth}
      \centering
      \includegraphics[width=\linewidth]{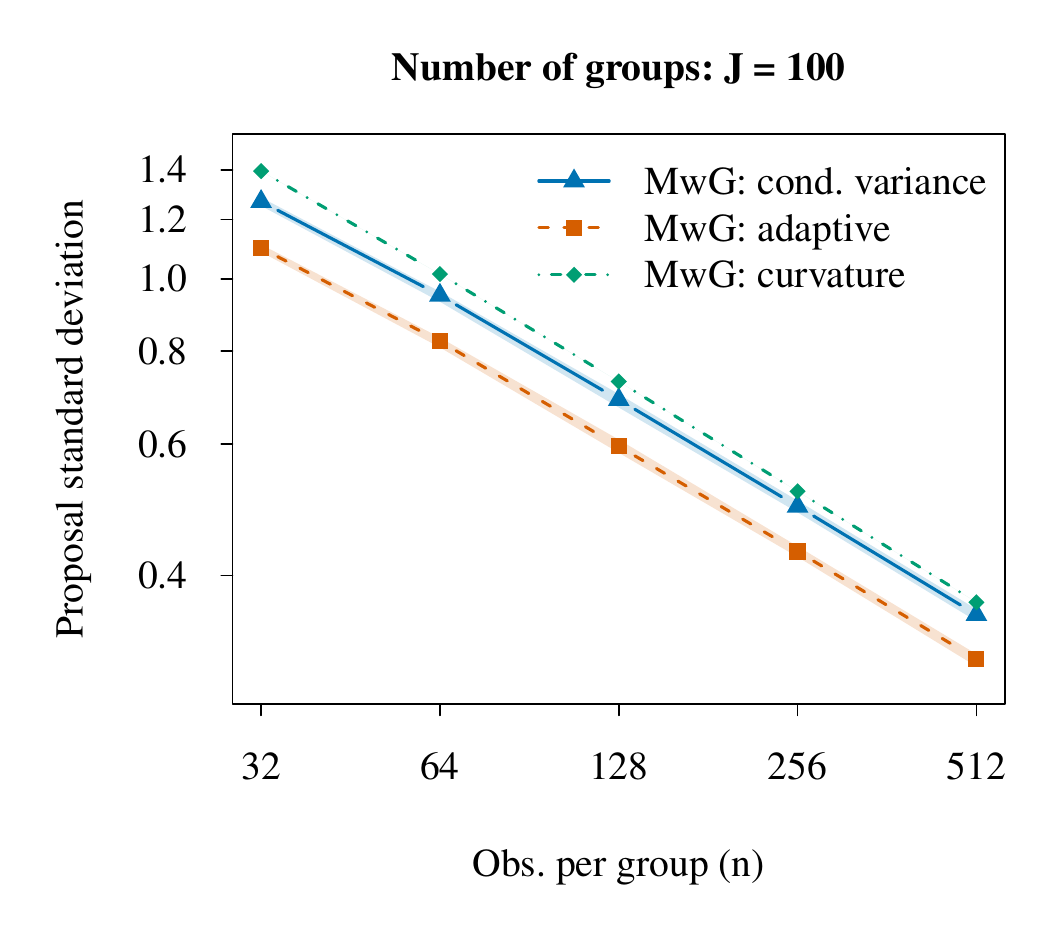}
    \end{subfigure}\hfill
    \begin{subfigure}[t]{0.5\linewidth}
      \centering
      \includegraphics[width=\linewidth]{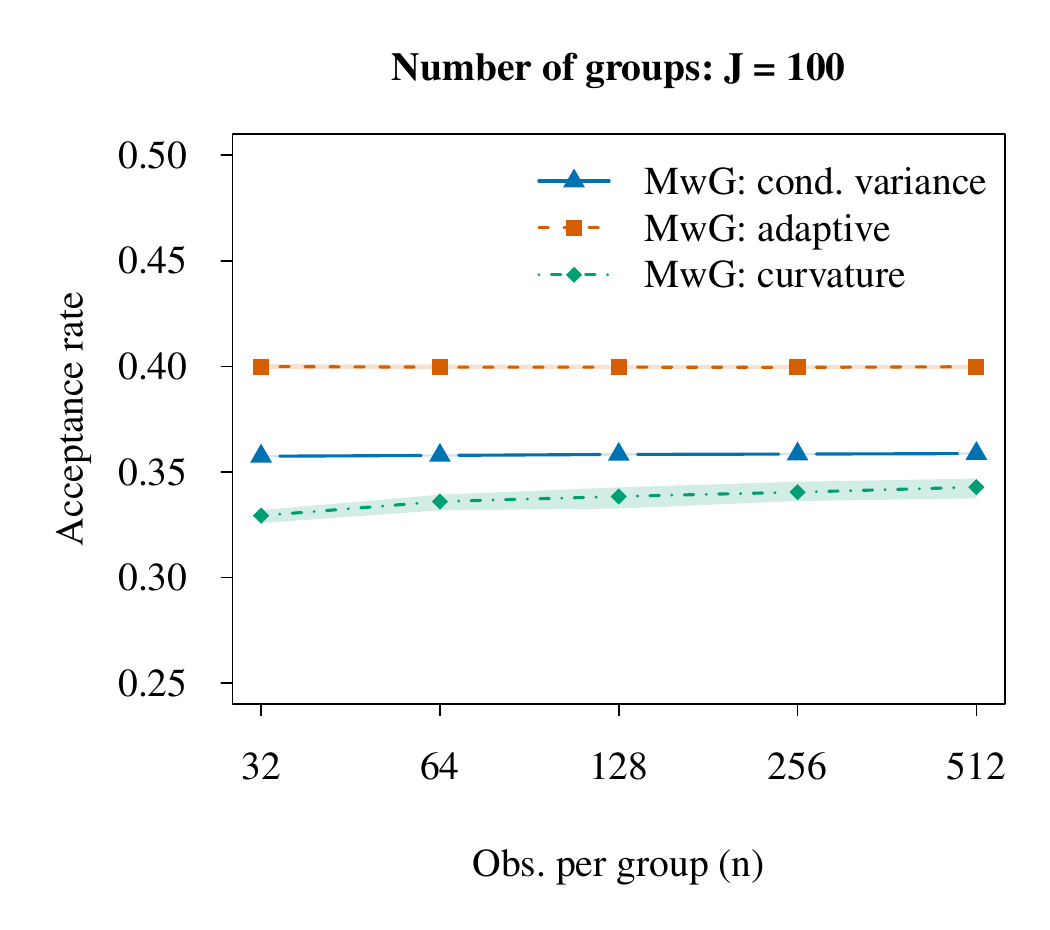}
    \end{subfigure}
  \end{minipage}\hfill
  \caption{Proposal standard deviations (left column) and acceptance rates (right column) for $J=10$ (top row) and $J=100$ (bottom row). Lines and points represent medians across the $100$ replicated datasets, while the shaded regions represent interquartile ranges.}
  \label{fig:standard-deviation-acceptance}
\end{figure}

Figure \ref{fig:standard-deviation-acceptance} reports the proposal standard deviations and acceptance rates of the three MwG schemes. For each replicated dataset, the reported proposal standard deviation and acceptance rate are averaged across coordinates and post-burn-in iterations. The curves show the median of these quantities across datasets, and the shaded regions show their interquartile ranges.

The proposal standard deviation of the curvature scheme decreases at a $n^{-1/2}$ rate and closely tracks the scale of the other proposal standard deviations. At the same time, the acceptance rates remain nearly constant across $n$, showing that the different methods are comparably tuned.

\bibliography{references}

@article{ascolani2025mixing,
  title={{Mixing times of data-augmentation Gibbs samplers for high-dimensional probit regression}},
  author={Ascolani, Filippo and Zanella, Giacomo},
  journal={Journal of Machine Learning Research},
  volume={27},
  number={127},
  pages={1--42},
  year={2026}
}

@article{kipnis1986central,
  title={{Central limit theorem for additive functionals of reversible Markov processes and applications to simple exclusions}},
  author={Kipnis, Claude and Varadhan, SR Srinivasa},
  journal={Communications in Mathematical Physics},
  volume={104},
  number={1},
  pages={1--19},
  year={1986},
  publisher={Springer}
}

@article{nesterov2012efficiency,
  title={{Efficiency of coordinate descent methods on huge-scale optimization problems}},
  author={Nesterov, Yu},
  journal={SIAM Journal on Optimization},
  volume={22},
  number={2},
  pages={341--362},
  year={2012},
  publisher={SIAM}
}

@article{bobkov1999isoperimetric,
  title={{Isoperimetric and analytic inequalities for log-concave probability measures}},
  author={Bobkov, Sergey G},
  journal={The Annals of Probability},
  volume={27},
  number={4},
  pages={1903--1921},
  year={1999},
  publisher={Institute of Mathematical Statistics}
}

@article{andrieu2024explicit,
  title={{Explicit convergence bounds for Metropolis Markov chains: Isoperimetry, spectral gaps and profiles}},
  author={Andrieu, Christophe and Lee, Anthony and Power, Sam and Wang, Andi Q},
  journal={The Annals of Applied Probability},
  volume={34},
  number={4},
  pages={4022--4071},
  year={2024},
  publisher={Institute of Mathematical Statistics}
}

@article{ascolani2024entropy,
  title={{Entropy contraction of the Gibbs sampler under log-concavity}},
  author={Ascolani, Filippo and Lavenant, Hugo and Zanella, Giacomo},
  journal={arXiv preprint arXiv:2410.00858},
  year={2024},
  note={Accepted for publication in The Annals of Probability}
}

@article{lee2024fast,
  title={{Fast Mixing of Data Augmentation Algorithms: Bayesian Probit, Logit, and Lasso Regression}},
  author={Lee, Holden and Zhang, Kexin},
  journal={arXiv preprint arXiv:2412.07999},
  year={2024},
    note={Accepted for publication in The Annals of Statistics}
}

@article{ascolani2023clustering,
  title={{Clustering consistency with Dirichlet process mixtures}},
  author={Ascolani, Filippo and Lijoi, Antonio and Rebaudo, Giovanni and Zanella, Giacomo},
  journal={Biometrika},
  volume={110},
  number={2},
  pages={551--558},
  year={2023},
  publisher={Oxford University Press}
}

@article{ascolani2026scalability,
  title={{Scalability of Metropolis-within-Gibbs schemes for high-dimensional Bayesian models}},
  author={Ascolani, Filippo and Roberts, Gareth O and Zanella, Giacomo},
  journal={Journal of the Royal Statistical Society Series B: Statistical Methodology},
  pages={qkaf084},
  year={2026},
  publisher={Oxford University Press UK}
}

@article{roberts1998two,
  title={{Two convergence properties of hybrid samplers}},
  author={Roberts, Gareth O and Rosenthal, Jeffrey S},
  journal={The Annals of Applied Probability},
  volume={8},
  number={2},
  pages={397--407},
  year={1998},
  publisher={Institute of Mathematical Statistics}
}

@article{tong2020mala,
  title={{MALA-within-Gibbs samplers for high-dimensional distributions with sparse conditional structure}},
  author={Tong, Xin T and Morzfeld, Matthias and Marzouk, Youssef M},
  journal={SIAM Journal on Scientific Computing},
  volume={42},
  number={3},
  pages={A1765--A1788},
  year={2020},
  publisher={SIAM}
}

@inproceedings{ding2021randomU,
  title={{Random coordinate underdamped langevin monte carlo}},
  author={Ding, Zhiyan and Li, Qin and Lu, Jianfeng and Wright, Stephen},
  booktitle={International conference on artificial intelligence and statistics},
  pages={2701--2709},
  year={2021},
  organization={PMLR}
}

@inproceedings{ding2021random,
  title={{Random coordinate langevin monte carlo}},
  author={Ding, Zhiyan and Li, Qin and Lu, Jianfeng and Wright, Stephen J},
  booktitle={Conference on learning theory},
  pages={1683--1710},
  year={2021},
  organization={PMLR}
}

@article{qin2024spectral,
  title={{Spectral telescope: Convergence rate bounds for random-scan Gibbs samplers based on a hierarchical structure}},
  author={Qin, Qian and Wang, Guanyang},
  journal={The Annals of Applied Probability},
  volume={34},
  number={1B},
  pages={1319--1349},
  year={2024},
  publisher={Institute of Mathematical Statistics}
}

@article{qin2025spectral2,
  title={{On spectral gap decomposition for Markov chains}},
  author={Qin, Qian},
  journal={arXiv preprint arXiv:2504.01247},
  year={2025}
}

@article{qin2025spectral,
  title={{Spectral gap bounds for reversible hybrid Gibbs chains}},
  author={Qin, Qian and Ju, Nianqiao and Wang, Guanyang},
  journal={The Annals of Statistics},
  volume={53},
  number={4},
  pages={1613--1638},
  year={2025},
  publisher={Institute of Mathematical Statistics}
}

@article{casella1992explaining,
  title={{Explaining the Gibbs sampler}},
  author={Casella, George and George, Edward I},
  journal={The American Statistician},
  volume={46},
  number={3},
  pages={167--174},
  year={1992},
  publisher={Taylor \& Francis}
}

@book{brooks2011handbook,
  title={{Handbook of markov chain monte carlo}},
  author={Brooks, Steve and Gelman, Andrew and Jones, Galin and Meng, Xiao-Li},
  year={2011},
  publisher={CRC press}
}

@article{chib1995understanding,
  title={{Understanding the metropolis-hastings algorithm}},
  author={Chib, Siddhartha and Greenberg, Edward},
  journal={The american statistician},
  volume={49},
  number={4},
  pages={327--335},
  year={1995},
  publisher={Taylor \& Francis}
}

@article{fort2003geometric,
  title={{On the geometric ergodicity of hybrid samplers}},
  author={Fort, G and Moulines, E and Roberts, Gareth O and Rosenthal, JS},
  journal={Journal of Applied Probability},
  volume={40},
  number={1},
  pages={123--146},
  year={2003},
  publisher={Cambridge University Press}
}

@article{qin2022convergence,
  title={{Convergence rates of two-component MCMC samplers}},
  author={Qin, Qian and Jones, Galin L},
  journal={Bernoulli},
  volume={28},
  number={2},
  pages={859--885},
  year={2022},
  publisher={Bernoulli Society for Mathematical Statistics and Probability}
}

@article{chen2020fast,
  title={{Fast mixing of Metropolized Hamiltonian Monte Carlo: Benefits of multi-step gradients}},
  author={Chen, Yuansi and Dwivedi, Raaz and Wainwright, Martin J and Yu, Bin},
  journal={Journal of Machine Learning Research},
  volume={21},
  number={92},
  pages={1--72},
  year={2020}
}

@article{amit1996convergence,
  title={{Convergence properties of the Gibbs sampler for perturbations of Gaussians}},
  author={Amit, Yali},
  journal={The Annals of Statistics},
  volume={24},
  number={1},
  pages={122--140},
  year={1996},
  publisher={Institute of Mathematical Statistics}
}

@article{baxendale2005renewal,
  title={{Renewal theory and computable convergence rates for geometrically ergodic Markov chains}},
  author={Baxendale, Peter H},
  journal={The Annals of Applied Probability},
  volume={15},
  number={1B},
  pages={700--738},
  year={2005}
}

@article{dwivedi2019log,
  title={{Log-concave sampling: Metropolis-Hastings algorithms are fast}},
  author={Dwivedi, Raaz and Chen, Yuansi and Wainwright, Martin J and Yu, Bin},
  journal={Journal of Machine Learning Research},
  volume={20},
  number={183},
  pages={1--42},
  year={2019}
}

@article{luu2024gibbs,
  title={{Is Gibbs sampling faster than Hamiltonian Monte Carlo on GLMs?}},
  author={Luu, Son and Xu, Zuheng and Surjanovic, Nikola and Biron-Lattes, Miguel and Campbell, Trevor and Bouchard-Cote, Alexandre},
  journal={Proceedings of the 28th International Conference on Artificial Intelligence and Statistics},
  series={Proceedings of Machine Learning Research},
  volume={258},
  pages={2881--2889},
  year={2025},
  publisher={PMLR}
}

@book{helmberg2008introduction,
  title={{Introduction to spectral theory in Hilbert space}},
  author={Helmberg, Gilbert},
  year={2008},
  publisher={Courier Dover Publications}
}

@book{gelman1995bayesian,
  title={{Bayesian data analysis}},
  author={Gelman, Andrew and Carlin, John B and Stern, Hal S and Rubin, Donald B},
  year={1995},
  publisher={Chapman and Hall/CRC}
}

@article{rosenthal2003asymptotic,
  title={{Asymptotic variance and convergence rates of nearly-periodic Markov chain Monte Carlo algorithms}},
  author={Rosenthal, Jeffrey S},
  journal={Journal of the American Statistical Association},
  volume={98},
  number={461},
  pages={169--177},
  year={2003},
  publisher={Taylor \& Francis}
}

@article{andrieu2008tutorial,
  title={{A tutorial on adaptive MCMC}},
  author={Andrieu, Christophe and Thoms, Johannes},
  journal={Statistics and computing},
  volume={18},
  number={4},
  pages={343--373},
  year={2008},
  publisher={Springer}
}

@article{gong2014practical,
  title={{A practical sequential stopping rule for high-dimensional Markov chain Monte Carlo}},
  author={Gong, Lei and Flegal, James M},
  journal={Journal of Computational and Graphical Statistics},
  volume={25},
  number={3},
  pages={684--700},
  year={2016},
  publisher={Taylor \& Francis}
}

@incollection{sokal1997monte,
  title={{Monte Carlo methods in statistical mechanics: foundations and new algorithms}},
  author={Sokal, Alan},
  booktitle={Functional integration: Basics and applications},
  pages={131--192},
  year={1997},
  publisher={Springer}
}

@article{lawler1988bounds,
  title={{Bounds on the $L^2$ spectrum for Markov chains and Markov processes: a generalization of Cheeger's inequality}},
  author={Lawler, Gregory F and Sokal, Alan D},
  journal={Transactions of the American mathematical society},
  volume={309},
  number={2},
  pages={557--580},
  year={1988}
}

@article{gilks1992adaptive,
  title={{Adaptive rejection sampling for Gibbs sampling}},
  author={Gilks, Walter R and Wild, Pascal},
  journal={Journal of the Royal Statistical Society: Series C (Applied Statistics)},
  volume={41},
  number={2},
  pages={337--348},
  year={1992},
  publisher={Wiley Online Library}
}

\end{document}